\documentclass[letterpaper]{article} 
\usepackage{aaai24}  
\usepackage{times}  
\usepackage{helvet}  
\usepackage{courier}  
\usepackage[hyphens]{url}  
\usepackage{graphicx} 
\usepackage{natbib}  
\usepackage{caption} 
\usepackage{algorithm}
\usepackage{algorithmic}

\usepackage{amsfonts}
\usepackage{amsmath}
\usepackage{multirow}
\usepackage{newfloat}
\usepackage{listings}
\DeclareCaptionStyle{ruled}{labelfont=normalfont,labelsep=colon,strut=off} 
\floatstyle{ruled}
\newfloat{listing}{tb}{lst}{}
\floatname{listing}{Listing}
\title{Towards Balanced Alignment: Modal-Enhanced Semantic Modeling \\ for Video Moment Retrieval}
\author{
    Zhihang Liu\textsuperscript{\rm 1},
    Jun Li\textsuperscript{\rm 2},
    Hongtao Xie\textsuperscript{\rm 1}\thanks{Corresponding author},
    Pandeng Li\textsuperscript{\rm 1},\\
    Jiannan Ge\textsuperscript{\rm 1},
    Sun-Ao Liu\textsuperscript{\rm 1},
    Guoqing Jin\textsuperscript{\rm 2}
}
\affiliations{
    \textsuperscript{\rm 1} University of Science and Technology of China, Hefei, China \ 
    \textsuperscript{\rm 2} People's Daily Online\\

    \{liuzhihang, lpd, gejn, lsa1997\}@mail.ustc.edu.cn,
    htxie@ustc.edu.cn,
    \{lijun, jinguoqing\}@people.cn
}

\usepackage{bibentry}

\begin{document}

\maketitle

\begin{abstract}
Video Moment Retrieval (VMR) aims to retrieve temporal segments in untrimmed videos corresponding to a given language query by constructing cross-modal alignment strategies.
However, these existing strategies are often sub-optimal since they ignore the modality imbalance problem, \textit{i.e.}, the semantic richness inherent in videos far exceeds that of a given limited-length sentence. 
Therefore, in pursuit of better alignment, a natural idea is enhancing the video modality to filter out query-irrelevant semantics, and enhancing the text modality to capture more segment-relevant knowledge. 
In this paper, we introduce Modal-Enhanced Semantic Modeling (MESM), a novel framework for more balanced alignment through enhancing features at two levels.
First, we enhance the video modality at the frame-word level through word reconstruction. This strategy emphasizes the portions associated with query words in frame-level features while suppressing irrelevant parts. Therefore, the enhanced video contains less redundant semantics and is more balanced with the textual modality.
Second, we enhance the textual modality at the segment-sentence level by learning complementary knowledge from context sentences and ground-truth segments. With the knowledge added to the query, the textual modality thus maintains more meaningful semantics and is more balanced with the video modality.
By implementing two levels of MESM, the semantic information from both modalities is more balanced to align, thereby bridging the modality gap.
Experiments on three widely used benchmarks, including the out-of-distribution settings, show that the proposed framework achieves a new start-of-the-art performance with notable generalization ability (\textit{e.g.}, 4.42\% and 7.69\% average gains of R1@0.7 on Charades-STA and Charades-CG). The code will be available at https://github.com/lntzm/MESM.
\end{abstract}

\section{Introduction}
Video moment retrieval (VMR) poses a meaningful and challenging task in video understanding. Given a natural language query that describes a moment segment in an untrimmed video, VMR aims to determine the start and end timestamps of the segment in the video~\cite{anne2017localizing, gao2017tall}. Therefore, it necessitates an accurate understanding of both the video content and the language query, as well as their alignment~\cite{li2023progressive}.

Modality alignment in existing VMR methods is primarily implemented at two distinct levels. Some previous studies~\cite{li2021proposal, liu2022memory} align frame-level and word-level features, devising efficient alignment strategies to accurately regress the moments. Another line of methods~\cite{chen2019semantic, wang2022negative} generates proposals to extract segment-level features, aligning them with sentence-level features to identify the most matching segment as the answer. There are also some methods considering both the frame-word and segment-sentence level~\cite{wang2021structured, moon2023query}. Normally, they first align frame-level and word-level features, then pool the segments for further alignment and moment retrieval.

\begin{figure}[t]
\centering
\includegraphics[width=0.47\textwidth]{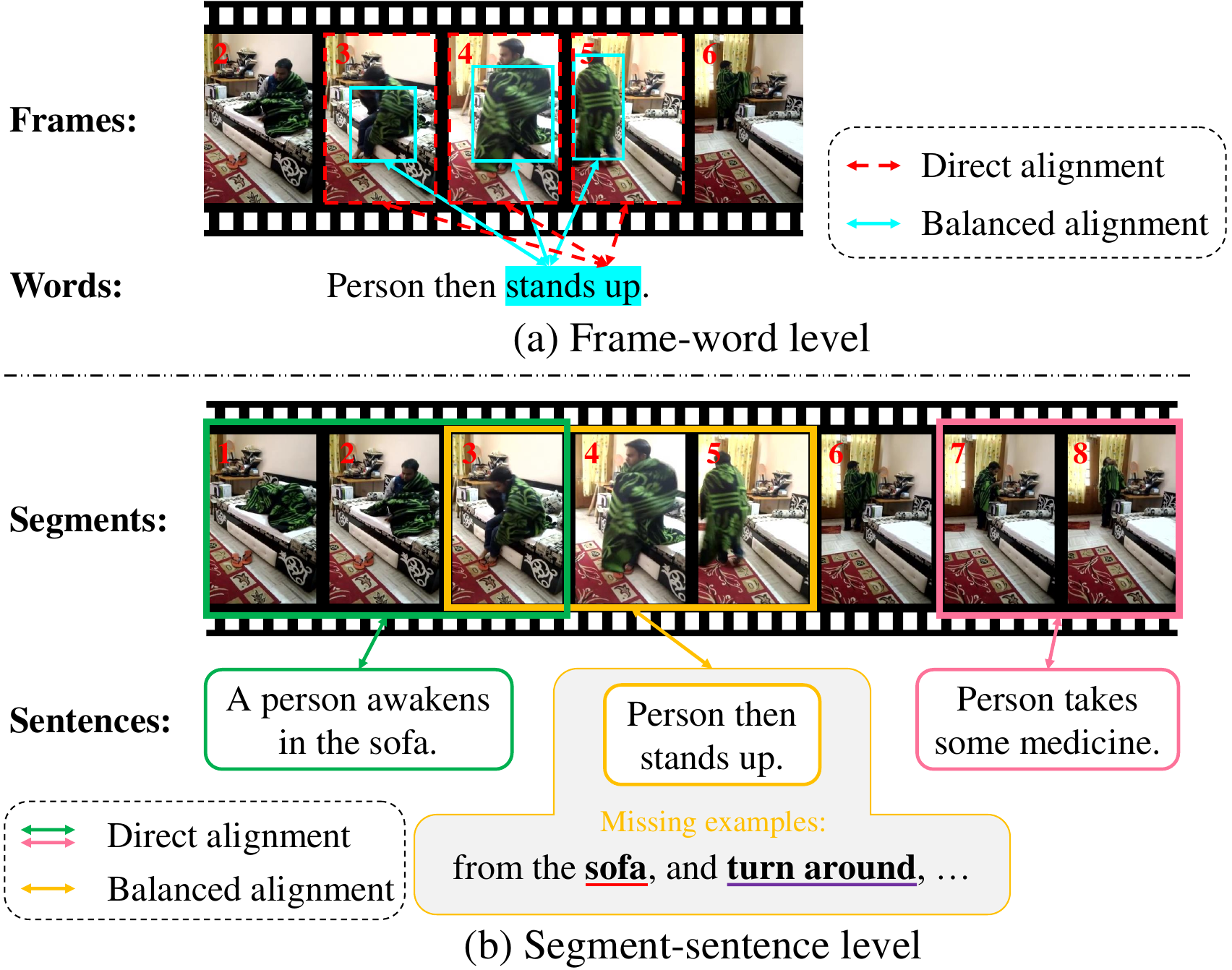}
\caption{We creatively analyze the modality imbalance problem in VMR and the comparison between existing direct alignment and our balanced alignment, which manifests in two levels: (a) Frame-word level, the description of a word should typically align with specific parts within a frame (balanced) rather than the entire frame (direct). (b) Segment-sentence level, there is some semantic information in the segment but absent in the given sentence. The segment should typically align with the expanded sentence semantics (balanced) rather than the sentence only (direct).}
\label{fig_intro_problem}
\end{figure}

Despite the achievements of existing alignment strategies, most of them disregard a crucial modality imbalance problem at both the frame-word and segment-sentence levels, resulting in a modality gap. 
At the frame-word level, as shown in Figure~\ref{fig_intro_problem}(a), the words in a sentence are typically aligned with specific parts within a frame rather than the entire frame (\textit{e.g.}, the action \textit{stands up}), which poses difficulties to understand the fine-grained relationship between two modalities.
Figure~\ref{fig_intro_problem}(b) shows the case of the segment-sentence level. First, the semantic information of the segment (\textit{e.g.}, frame \#3 to \#5) surpasses the details provided in the given sentence and humans can easily infer missing information (\textit{e.g.}, \textit{from the sofa}). Second, the sentence itself may be ambiguous for VMR due to the annotation subjectivity. For example, frame \#5 captures the action of \textit{turn around}, which is entirely absent in the given sentence. Both scenarios result in a negative impact on video understanding. In summary, due to the inherent semantic richness of the video modality, the textual modality should only align to a subset of video modality at both levels, and direct alignment with the entire modalities thus results in sub-optimal solutions.

To tackle the problem, a natural idea is to enhance both modalities simultaneously. The video modality should be enhanced to filter out irrelevant semantics for the query, and the textual modality should be enhanced to capture more knowledge related to the segment. Therefore, we propose a novel framework named Modal-Enhanced Semantic Modeling (MESM) to enhance them at two levels.
At the frame-word level, we enhance the video modality by reconstructing words via a weight-shared cross-attention mechanism. Since the words typically refer to certain portions of the frames, the reconstruction renders the model more sensitive to these semantically relevant portions and suppresses irrelevant ones. Consequently, there is less redundant semantic information in the output enhanced video feature, thus more balanced with words. 
At the segment-sentence level, we enhance the textual modality by learning complementary knowledge for the given query. As shown in Figure~\ref{fig_intro_problem}(b), the absence of semantic knowledge typically originates from both the given sentences within a video (\textit{e.g.}, \textit{sofa}, underlined in red) and the scenes of video segments (\textit{e.g.}, \textit{turn around}, underlined in purple).
Therefore, we can acquire the absent semantics by learning from both sources. We mask the given sentence and regenerate the semantic knowledge supervised by the corresponding segment. As the generated knowledge complements the query, the semantic information of the query becomes stronger and is thus more balanced with the segment.
Extensive experiments show our MESM achieves new state-of-the-art performance on three benchmarks and the out-of-distribution settings, demonstrating improved modality alignment and generalization.

The main contributions of our paper can be listed as follows. (1) As far as we know, we are the first to analyze the modality imbalance problem in VMR from both the frame-word and segment-sentence levels. (2) To alleviate the modality imbalance problem, we propose a novel framework MESM to model the enhanced semantic information from two levels, balancing the alignment to bridge the modality gap. (3) Extensive experimental results demonstrate the effectiveness of the proposed method.

\section{Related Work}

\noindent\textbf{Video Moment Retrieval.}
Different from Video Retrieval~\cite{li2022dual}, Video Moment Retrieval is a cross-modal task that emphasizes the ability to understand both video and textual modalities, including their alignment. The alignment can be typically split into the frame-word and segment-sentence levels. Some methods align the frame-level feature with the word-level feature~\cite{yuan2019find, zhang2020span, liu2021context, li2021proposal, liu2022memory}. Normally, they design various alignment strategies to directly predict the start and end moments.
Other methods focus on aligning the segment-level feature with the sentence-level feature~\cite{gao2017tall, chen2019semantic, zhang2020learning, wang2022negative}. They usually generate proposals to obtain the segment-level feature and align it with the sentence-level feature to select the best matching segment.
There are recently some methods implementing the alignment from both levels~\cite{wang2021structured, sun2022you, moon2023query, wang2023ms}. SMIN~\cite{wang2021structured} carefully designs multi-level alignment based on 2D-TAN. DETR-based methods~\cite{lei2021detecting, moon2023query, wang2023ms, momentdiff} usually do the frame-word level alignment, and then pool the segments with learnable proposals for further interaction, which yields promising results. However, most of these methods overlook the modality imbalance problem, leading to the modality gap.

\noindent\textbf{Modality Imbalance Problem.}
The modality imbalance problem seems widely existing in video-text representation tasks. \cite{ko2022video} points out the non-sequential alignment problem between the video and the text due to the ambiguity of labeling and designed a differentiable weak temporal alignment. \cite{wu2023cap4video} used large language models to generate auxiliary captions for a video to complete the video-text retrieval task.
In VMR, the modality imbalance problem is also crucial but few researchers focus on it. \cite{ding2021support} builds a support set using generative captions, considering the co-existence of some visual entities. Still, many methods only use one video-query pair as their input, ignoring the causal relationship among different sentences of segments within the same video and simply considering these sentences as negative ones~\cite{wang2022negative, luo2023towards}. Different from them, we utilize this information with the video modality together, enhancing both the video and textual modalities, leading to a more balanced alignment and bridging the modality gap.

\section{Proposed Method}

\begin{figure*}[t]
\centering
\includegraphics[width=0.91\textwidth]{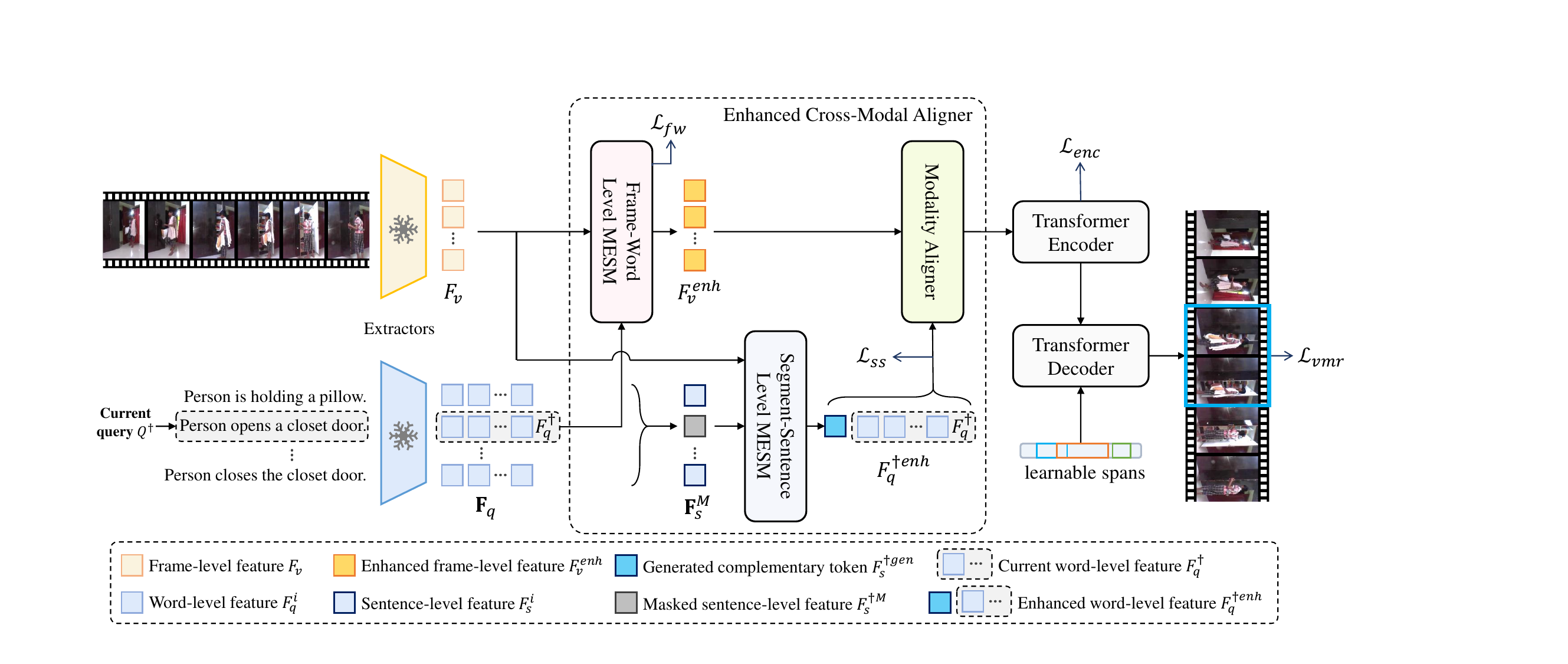}
\caption{An overview of our MESM, which includes the feature extractors, the proposed Enhanced Cross-Modal Aligner (ECMA), and a transformer encoder-decoder network. We model the enhanced semantic information at two levels in ECMA, which consists of the Frame-Word level MESM (FW-MESM) and the Segment-Sentence level MESM (SS-MESM).}
\label{fig_pipeline}
\end{figure*}

\subsection{Overview}

\noindent\textbf{Problem Formulation.}
Given a pair of an untrimmed video $V=\{f_i\}_{i=1}^{N_v}$ and a language query $Q^\dag=\{w_i^\dag\}_{i=1}^{N_w}$, VMR aims to predict a video segment of moment $\hat{m}=(\hat{t_s}, \hat{t_e})$ that is most relevant to $Q^\dag$, where $N_v$ and $N_w$ represent the number of frames and words, respectively, $\hat{t_s}$ and $\hat{t_e}$ indicate the predicted start and end time of the video segment. 

\noindent\textbf{Pipeline.}
Figure~\ref{fig_pipeline} shows the pipeline of the proposed MESM, which consists of three steps. First, an offline video and text feature extractor is utilized to obtain frame-level and word-level features. Then, we design an Enhanced Cross-Modal Aligner (ECMA) to alleviate the modality imbalance problem and complete a more balanced alignment. Last, a transformer encoder-decoder network is utilized to encode the aligned feature and decode the moments from learnable spans.
Different from many methods that directly align the features of different modalities, we focus on balancing the alignment through modal-enhanced semantic modeling from both frame-word and segment-sentence levels in the proposed ECMA, bridging the modality gap.

\subsection{Feature Extractors}
Feature extractors are necessary for downstream tasks~\cite{du2022svtr, zheng2023cdistnet, zhang2023linguistic}.
Followed by most VMR methods~\cite{zhang2020learning, wang2023ms}, we use offline feature extractors to get pre-obtained features from the raw data of the video and text.
Generally, given a video extractor and a text extractor, we use trainable MLPs to map the extracted video feature and text feature to a common space. 
Given a set of language queries $\mathbf{Q} = \{Q^i | i=1, ..., K\}$ belonging to the same video $V$, the mapped video and text feature can be represented as $F_v \in \mathbb{R}^{L_v \times D}$ and $\mathbf{F}_q = \{F_q^i \in \mathbb{R}^{L_w \times D} | i=1, ..., K\}$, respectively, where $K$ is the number of sentences in the video, $D$ is the dimension of the common space. $L_v$ and $L_w$ are the lengths of the features.
We use $F_q^\dag \in \mathbf{F}_q$ to represent the feature of current query $Q^\dag$ for moment retrieval, and thus $F_v$ and $F_q^\dag$ are frame-level and word-level features, respectively.

\subsection{Enhanced Cross-Modal Aligner}
This section presents our proposed Enhanced Cross-Modal Aligner, comprising three sub-modules: Frame-Word level MESM (FW-MESM), Segment-Sentence level MESM (SS-MESM), and the Modality Aligner (MA). FW-MESM enhances the video modality at the frame-word level by emphasizing the query-relevant portions of frame-level features
and suppressing irrelevant ones.
SS-MESM enhances the text modality at the segment-sentence level by generating a complementary token derived from both the query set and the ground-truth segment.
Given that FW-MESM and SS-MESM generate enhanced features to address the modality imbalance issue, we subsequently implement MA to achieve the ultimate cross-modal alignment.

\begin{figure}[t]
\centering
\includegraphics[width=0.4\textwidth]{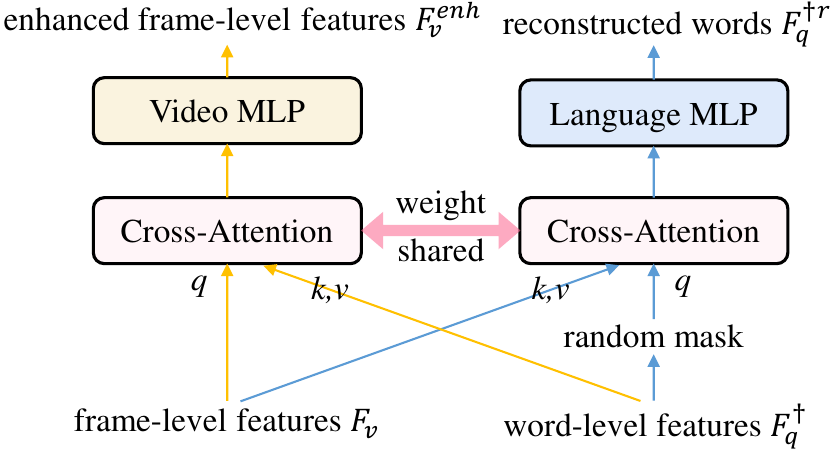}
\caption{The pipeline of FW-MESM. The weights of the cross-attention are shared, inputs are exchanged for MLM.}
\label{fig_FWMLM}
\end{figure}

\noindent\textbf{Frame-Word Level MESM.}
Since words often refer to specific parts within frames~\cite{ge2021semantic, ge2022dual}, we enhance the frame-level feature to filter out redundant parts and design an efficient semantic modeling strategy based on a weight-shared cross-attention mechanism.
It is proven that weight-shared self-attention can process data from different modalities~\cite{bao2022vlmo, wang2023image}, and we expand it to the case of cross-attention. 
As shown in Figure~\ref{fig_FWMLM}, the output of cross-attention $F_v^{enh}$ (the left branch) symbolizes the video feature targeted for enhancement. The enhancement necessitates the acquisition of fine-grained discrimination ability to emphasize word-relevant parts within the frames, and we implement it by an auxiliary masked language modeling (MLM) task with the weight-shared cross attention (the right branch). Once the ability is obtained, the shared weights provide a bridge to enhance the output.

Specifically, we first treat the projection of frame-level feature $\mathcal{Q}=W^v F_v$ as \textit{query}, the projection of word-level feature $\mathcal{K} = W^k F_q^\dag$ and $\mathcal{V} = W^v F_q^\dag$ as \textit{key} and \textit{value}, where $W^q$, $W^k$, $W^v$ are linear projection matrices. Therefore, the output feature $F_v^{enh} \in \mathbb{R}^{L_v \times D}$ can be formulated as:
\begin{equation}
    F_v^{enh} = F_v + \text{MLP}\left(
        \text{softmax}\left(
            \frac{\mathcal{Q}\mathcal{K}^\top}{\sqrt{d}}
        \right)\mathcal{V}
    \right),
\label{eq_cross-attn}
\end{equation}
where $d$ is the dimension of the \textit{query}, \textit{key} and \textit{value}. To enhance $F_v^{enh}$, we exchange the modality of the input and employ MLM. During the MLM, 1/3 of the words are randomly masked. 
If we denote the masked word-level feature as $F_q^{\dag m}$, and the modal-exchanged inputs are $\mathcal{Q}^* = W^q F_q^{\dag m}$ for \textit{query}, $\mathcal{K}^* = W^k F_v$ and $\mathcal{V}^* = W^v F_v$ for \textit{key} and \textit{value}, the reconstructed feature of words $F_q^{\dag r} \in \mathbb{R}^{L_w \times D}$ can be calculated similar to Equation~\ref{eq_cross-attn}.
Then a fully connected layer and softmax operation is utilized to get the probability distribution $P(F_q^{\dag r}) \in \mathbb{R}^{L_w \times N_{\text{vocab}}}$ of the words, where $N_{\text{vocab}}$ is the vocabulary size. We use the cross-entropy loss to measure the similarity between the reconstructed words and the original words, which can be formulated as:
\begin{equation}
    \mathcal{L}_{fw} = - \frac{1}{L_w} \sum_{j=1}^{L_w}{
        z_j^\dag \log P_j(F_q^{\dag r})
    },
\end{equation}
where $z_j$ is the label of the $j$-th word in a sentence.

Due to the shared weights, the obtained ability from the MLM task applies to the original output as well. Therefore, $F_v^{enh}$ is enhanced to highlight the semantically relevant portions in $F_v$ while filtering out irrelevant portions, making it more balanced with the textual modality.

\noindent\textbf{Segment-Sentence Level MESM.}
Since the sentence can not fully cover the segment,  
we enhance the textual modality at the segment-sentence level by generating a complementary token from context sentences and the ground-truth segment, then the token is concatenated to the given query. To supervise the learning of the complementary knowledge, we construct a positive set for contrastive learning. The positive set collects the existing neighborhoods of the ground-truth segment to perform a soft supervision since they own similar semantic information, which we introduce later.

As we have extracted the word-level feature $\mathbf{F}_q$ of the $K$ sentences, 
we simply average $F_q^i$ for $i$-th sentence to get the sentence-level feature $F_s^i = \frac{1}{L_w} \sum_{j=1}^{L_w}{(F_q^i)_j}$ and thus $F_s^i \in \mathbb{R}^{D}$.
For the current sentence-level feature $F_s^\dag$, we replace it with a learnable [MASK] token $F_s^{\dag M} \in \mathbb{R}^{D}$, and the set of the sentence-level feature with the masked one can be written as $\mathbf{F}_s^M = \{ F_s^1, ..., F_s^{\dag M},..., F_s^K\} \in \mathbb{R}^{K \times D}$. Then the cross-attention layers are implemented on $\mathbf{F}_s^M$ (\textit{query}) and $F_v$ (\textit{key} and \textit{value}).
The output of the cross-attention layers can be represented as:
\begin{equation}
    \mathbf{F}_s^{gen} = \{ 
        F_s^{1gen},...,F_s^{\dag gen},...,F_s^{Kgen}
    \} \in \mathbb{R}^{K \times D}.
\end{equation}
Note that $F_s^{\dag gen}$ is in the output $\mathbf{F}_s^{gen}$ and we take it out as the generated token for the complementary knowledge to the sentence.
Then we concatenate the generated token $F_s^{\dag gen}$ with the word-level feature $F_q^\dag$ together to get the enhanced word-level feature $F_q^{\dag enh}  = [F_s^{\dag gen}, F_q^\dag] \in \mathbb{R}^{(L_w + 1) \times D}$.

With the complementary knowledge, the query is more balanced with the segment. Thus, we use the segment-level feature to supervise $F_q^{\dag enh}$ to obtain knowledge related to the segment. Given the ground truth of the video segment $(l_s, l_e)$, where $l_s$ and $l_e$ denote the start and the end index of the frame-level feature $F_v$. We take the average as the segment-level feature $S \in \mathbb{R}^{D}$, which can be formulated as:
\begin{equation}
    S = \frac{1}{l_e+1-l_s}\sum_{j=l_s}^{l_e}{
        (F_v)_j
    }.
\end{equation}
Then we design a contrastive loss to supervise the knowledge learning. Since there may be some neighbor segments with similar moments in the video, we build a positive set $S_{pos}$ in a batch based on the IoU among the segments. We take the segments as positive when the IoU between two of them is larger than $\gamma$, 
and the corresponding knowledge should be similar. The contrastive loss can be formulated as:
\begin{equation}
    \mathcal{L}_{ss} = - \log \frac{
        \sum_{j \in S_{pos}}{\exp(\sum_{k=1}^{L_w+1}{F_q^{\dag enh}} \cdot S / \tau)}
    }{
        \sum_{j=1}^{N_b}{\exp(\sum_{k=1}^{L_w+1}{F_q^{\dag enh}} \cdot S / \tau)}
    },
\end{equation}
where $N_b$ denotes the batch size, $\tau$ is the temperature coefficient. Supervised by the segment-level feature $S$, the enhanced word-level $F_s^{\dag enh}$ thus contains the complementary semantic information within the whole segment and is thus more balanced with the video modality.

\noindent\textbf{Modality Aligner.}
Since we have gotten the enhanced frame-level feature $F_v^{enh}$ and enhanced word-level feature $F_q^{\dag enh}$, we finally employ cross-attention layers between $F_v^{enh}$ (\textit{query}) and $F_q^{\dag enh}$ ( \textit{key} and \textit{value}) to do the modality interaction and alignment. 
The final aligned feature $F \in \mathbb{R}^{L_v \times D}$ can be calculated as the standard cross-attention.

\subsection{Transformer Encoder-Decoder}
After the modal-aligned feature $F$ is obtained from ECMA, a DETR~\cite{carion2020end} network is utilized to complete the VMR, which consists of a transformer encoder and decoder.
The transformer encoder encodes $F$ to a fusion representation $F_{enc}$, helping the model better understand the sequence relations. The encoding process follows the standard self-attention and the loss can be calculated as:
\begin{equation}
    \mathcal{L}_{enc} = - \frac{1}{L_v} \sum_{j=1}^{L_v}{
        y_j \log(s_j) + (1 - y_j) \log(1 - s_j)
    },
\end{equation}
where $s \in \mathbb{R}^{L_v}$ is the similarity vector which represents the attention of the model to focus, and is obtained by an MLP from $F_{enc}$. $y \in \mathbb{R}^{L_v}$ is the similarity label, where $y_j$=$1$ if the $j$-th frame is within the ground-truth and $y_j$=$0$ otherwise.

As for the transformer decoder, inspired by DAB-DETR~\cite{liu2021dab}, we follow QD-DETR~\cite{moon2023query} to design learnable spans, representing the center coordinate and the window. The decoder calculates the standard cross-attention between learnable spans and the pooled features, refining the result of spans continually.

Inspired by~\cite{carion2020end, lei2021detecting}, the moment retrieval loss consists of three parts:
\begin{equation}
    \mathcal{L}_{vmr} = \lambda_{L1}\| m - \hat{m} \|_1 + 
        \lambda_{iou} \mathcal{L}_{iou}(m, \hat{m}) +
        \lambda_{ce} \mathcal{L}_{ce},
\end{equation}
where $m$ and $\hat{m}$ are the predicted and ground-truth moments, $\lambda_{(L1, iou, ce)}$ are the hyper-parameters, $\mathcal{L}_{iou}$ is the generalized IoU loss~\cite{union2019metric}, $\mathcal{L}_{ce}$ is the cross-entropy loss to classify the foreground or background~\cite{carion2020end}.

As a result, the final loss is:
\begin{equation}
    \mathcal{L} = \lambda_{fw}\mathcal{L}_{fw} +
        \lambda_{ss}\mathcal{L}_{ss} +
        \lambda_{enc}\mathcal{L}_{enc} +
        \mathcal{L}_{vmr},
\end{equation}
where $\lambda_{fw}$, $\lambda_{ss}$ and $\lambda_{enc}$ are the hyper-parameters.

\begin{table}[t]
\centering
\resizebox{0.47\textwidth}{!}{%
\begin{tabular}{l|c|ccccc}
\hline
\multirow{3}{*}{Methods} & \multirow{3}{*}{Extractors}                                           & \multicolumn{5}{c}{Charades-STA}                                                               \\ \cline{3-7} 
                         &                                                                       & \multicolumn{2}{c|}{R1}                     & \multicolumn{3}{c}{mAP}                          \\
                         &                                                                       & @0.5           & \multicolumn{1}{c|}{@0.7}  & @0.5           & @0.75          & avg            \\ \hline
2D-TAN*                  & \multirow{8}{*}{\begin{tabular}[c]{@{}c@{}}VGG,\\ GloVe\end{tabular}} & 41.34          & \multicolumn{1}{c|}{23.91} & 54.68          & 24.15          & 29.26          \\
CBLN                     &                                                                       & 47.94          & \multicolumn{1}{c|}{28.22} & -              & -              & -              \\
RaNet*                   &                                                                       & 42.91          & \multicolumn{1}{c|}{25.82} & 53.28          & 24.41          & 28.55          \\
DCM                     &                                                                       & 47.80          & \multicolumn{1}{c|}{28.00} & -              & -              & -              \\
MMN*                     &                                                                       & 46.93          & \multicolumn{1}{c|}{27.07} & 58.85          & 28.16          & 31.58          \\
UMT\dag                  &                                                                       & 48.44          & \multicolumn{1}{c|}{29.76} & 58.03          & 27.46          & 30.37          \\
QD-DETR*                 &                                                                       & 51.51          & \multicolumn{1}{c|}{32.69} & 62.88          & 32.60          & 34.46          \\
MESM(Ours)               &                                                                       & \textbf{56.69} & \multicolumn{1}{c|}{\textbf{35.99}} & \textbf{67.94} & \textbf{33.64} & \textbf{37.33} \\ \hline
VDI                      & C, C                                                                  & 52.32          & \multicolumn{1}{c|}{31.37} & -              & -              & -              \\ \cline{2-7} 
M-DETR*                  & \multirow{3}{*}{C+SF, C}                                              & 53.22          & \multicolumn{1}{c|}{30.87} & 58.86          & 26.43          & 30.43          \\
QD-DETR*                 &                                                                       & 56.89          & \multicolumn{1}{c|}{32.50} & 66.49          & 32.00          & 35.39          \\
MESM(Ours)               &                                                                       & \textbf{61.24} & \multicolumn{1}{c|}{\textbf{38.04}} & \textbf{70.31} & \textbf{36.36} & \textbf{38.57} \\ \hline
\end{tabular}%
}
\caption{Performance comparison (\%) on the Charades-STA dataset. "\dag" means the method uses the audio data. "*" denotes that we re-implement the method under the same training scheme. M-DETR is short for MomentDETR.}
\label{tab:charades}
\end{table}

\begin{table}[!t]
\centering
\resizebox{0.45\textwidth}{!}{%
\begin{tabular}{l|cccc}
\hline
\multirow{2}{*}{Methods} & \multicolumn{4}{c}{TACoS}                             \\ \cline{2-5} 
                         & R1@0.1 & R1@0.3 & \multicolumn{1}{c|}{R1@0.5} & mIoU  \\ \hline
VSLNet                   & -      & 29.61  & \multicolumn{1}{c|}{24.27}  & 24.11 \\
2D-TAN                   & 47.59  & 37.29  & \multicolumn{1}{c|}{25.32}  & -     \\
CBLN                     & 49.16  & 38.98  & \multicolumn{1}{c|}{27.65}  & -     \\
RaNet                    & -      & 43.34  & \multicolumn{1}{c|}{33.54}  & -     \\
SeqPAN                   & -      & 31.72  & \multicolumn{1}{c|}{27.19}  & 25.86 \\
SMIN                     & -      & 48.01  & \multicolumn{1}{c|}{35.24}  & -     \\
MMN                      & 51.39  & 39.24  & \multicolumn{1}{c|}{26.17}  & -     \\
MS-DETR                  & -      & 47.66  & \multicolumn{1}{c|}{37.36}  & 35.09 \\
MESM(Ours)                & \textbf{65.03}  & \textbf{52.69}  & \multicolumn{1}{c|}{\textbf{39.52}}  & \textbf{36.94} \\ \hline
\end{tabular}%
}
\caption{Performance comparison (\%) on TACoS. All the listed methods use C3D and GloVe as their extractors.}
\label{tab:tacos}
\end{table}

\section{Experiments}

\begin{table}[t]
\centering
\resizebox{0.45\textwidth}{!}{%
\begin{tabular}{l|ccccc}
\hline
\multirow{3}{*}{Methods} & \multicolumn{5}{c}{QVHighlights}                                                                       \\ \cline{2-6} 
                         & \multicolumn{2}{c|}{R1}                             & \multicolumn{3}{c}{mAP}                          \\
                         & @0.5           & \multicolumn{1}{c|}{@0.7}          & @0.5           & @0.75          & avg         \\ \hline
MCN                      & 11.41          & \multicolumn{1}{c|}{2.72}          & 24.94          & 8.22           & 10.67          \\
CAL                      & 25.49          & \multicolumn{1}{c|}{11.54}         & 23.40          & 7.65           & 9.89           \\
XML                      & 41.83          & \multicolumn{1}{c|}{30.35}         & 44.63          & 31.73          & 32.14          \\
XML+                     & 46.69          & \multicolumn{1}{c|}{33.46}         & 47.89          & 34.67          & 34.90          \\
MomentDETR               & 52.89          & \multicolumn{1}{c|}{33.02}         & 54.82          & 29.40          & 30.73          \\
UMT\dag                  & 56.23          & \multicolumn{1}{c|}{41.18}         & 53.38          & 37.01          & 36.12          \\
QD-DETR                  & 62.40          & \multicolumn{1}{c|}{44.98}         & 62.52          & 39.88          & 39.86          \\
MESM(Ours)                & \textbf{62.78} & \multicolumn{1}{c|}{\textbf{45.20}}& \textbf{62.64} & \textbf{41.45} & \textbf{40.68} \\ \hline
\end{tabular}%
}
\caption{Performance comparison (\%) on QVHighlights \textit{test} split. All the listed methods use C+SF and C as their extractors. "\dag" denotes they use audio data.}
\label{tab:qvhighlights}
\end{table}

\begin{table*}[!t]
\centering
\resizebox{0.85\textwidth}{!}{%
\begin{tabular}{lcc|ccc|ccc}
\hline
\multirow{2}{*}{Methods} & \multirow{2}{*}{Year} & \multirow{2}{*}{Extractors} & \multicolumn{3}{c|}{Novel-composition}           & \multicolumn{3}{c}{Novel-word}                   \\ \cline{4-9} 
                         &                       &                             & R1@0.5         & R1@0.7         & mIoU           & R1@0.5         & R1@0.7         & mIoU           \\ \hline
LGI                      & 2020                  & I3D, GloVe                  & 29.42          & 12.73          & 30.09          & 26.48          & 12.47          & 27.62          \\
VSLNet                   & 2020                  & I3D, GloVe                  & 24.25          & 11.54          & 31.43          & 25.60          & 10.07          & 30.21          \\
VISA                     & 2022                  & I3D, GloVe                  & 45.41          & 22.71          & \textbf{42.03} & 42.35          & 20.88          & 40.18          \\
MESM(Ours)                & 2024                  & I3D, GloVe                  & \textbf{46.19} & \textbf{26.00} & 41.40          & \textbf{50.50} & \textbf{33.67} & \textbf{46.20} \\ \hline
MomentDETR*              & 2021                  & C+SF, C                     & 37.65          & 18.91          & 36.17          & 43.45          & 21.73          & 38.37          \\
QD-DETR*                 & 2023                  & C+SF, C                     & 40.62               & 19.96               & 36.64               & 48.2               & 26.19               & 43.22               \\
VDI                      & 2023                  & C, C                        & -              & -              & -              & 46.47          & 28.63          & 41.60          \\
MESM(Ours)                & 2024                  & C+SF, C                     & \textbf{44.39} & \textbf{23.27} & \textbf{39.89} & \textbf{52.66} & \textbf{31.22} & \textbf{46.38} \\ \hline
\end{tabular}%
}
\caption{Performance comparison (\%) on Charades-CG, which contains two types of OOD settings on Charades-STA: novel-composition and novel-word. "*" denotes the result we re-implement under the same training scheme.}
\label{tab:charades_cg}
\vspace{-4pt}
\end{table*}

\begin{table}[!t]
\centering
\resizebox{0.45\textwidth}{!}{
\begin{tabular}{ccc|cc|cc}
\hline
FW       & SS       & $\mathcal{L}_{enc}$ & R1@0.5         & R1@0.7         & mIoU           & mAP$_\text{avg}$      \\ \hline
           &            &                   & 53.82          & 30.78          & 46.75          & 33.99          \\
\checkmark &            &                   & 54.76          & 31.51          & 47.15          & 34.18          \\
           & \checkmark &                   & 54.60          & 33.17          & 47.60          & 34.24          \\
           &            & \checkmark        & 57.66          & 35.00          & 49.91          & 36.82          \\ \hline
\checkmark &            & \checkmark        & 59.62          & 36.26          & 50.91          & 37.83          \\
           & \checkmark & \checkmark        & 60.19          & 37.39          & 51.15          & 38.31          \\
\checkmark & \checkmark & \checkmark        & \textbf{61.24} & \textbf{38.04} & \textbf{52.14} & \textbf{38.57} \\ \hline
\end{tabular}
}
\caption{Main ablation study (\%) of module FW-MESM (FW) and SS-MESM (SS), loss $\mathcal{L}_{enc}$ on Charades-STA.}
\label{tab:ablation}
\vspace{-4pt}
\end{table}

\subsection{Experimental Settings}
\noindent\textbf{Datasets.}
We evaluate the proposed method on three widely used datasets, which are Charades-STA~\cite{gao2017tall}, TACoS\cite{regneri2013grounding}, and QVHighlights~\cite{lei2021detecting}. We also experiment on Charades-CG~\cite{li2022compositional}, which proposes out-of-distribution (OOD) settings for Charades-STA.
Charades-STA is built upon the Charades dataset~\cite{sigurdsson2016hollywood}, which consists of daily indoor activities. TACoS includes long-term videos about cooking activities. videos in QVHighlights range from daily vlog, travel vlog, and news. Charades-CG is proposed to evaluate the generalization ability by constructing new splits. These datasets cover videos from different domains, which are suitable for our evaluation in multiple scenes.

\noindent\textbf{Metrics.}
We calculate R1@$\mu$, mAP@$\mu$, mIoU, and mAP$_\text{avg}$ as used in previous methods~\cite{lei2021detecting, zhang2020learning, zhang2020span}. R1@$\mu$ and mAP@$\mu$ are the recall and mean average precision with IoU thresholds $\mu$ within the top-1 results. mIoU denotes the average IoU and mAP$_\text{avg}$ indicates the average mAP with $\mu$=[0.5:0.05:0.95].

\noindent\textbf{Implementation Details.}
We use different offline feature extractors for a fair comparison. VGG~\cite{simonyan2014very}, I3D~\cite{carreira2017quo}, C3D~\cite{tran2015learning} and C+SF (short for CLIP+SlowFast)~\cite{radford2021learning, feichtenhofer2019slowfast} are utilized as video extractors, GloVe~\cite{pennington2014glove} and C (short for CLIP) are used as text extractors. We set $\gamma$ as 0.9, the hidden dimension of the transformer layers as 256, the layers of FW-MESM, MA, transformer encoder, and decoder as 2. We build our model upon QD-DETR~\cite{moon2023query} with some optimizations, and train our model with Adam optimizer~\cite{kingma2014adam} on a single NVIDIA RTX 3090.

\subsection{Performance Comparisons}

\begin{table}[!t]
\centering
\resizebox{0.45\textwidth}{!}{%
\begin{tabular}{cc|cc|cc}
\hline
\multicolumn{2}{c|}{cross-attention layers}                                                                           & R1@0.5         & R1@0.7         & mIoU           & mAP$_\text{avg}$ \\ \hline
\multicolumn{1}{c|}{\multirow{3}{*}{\begin{tabular}[c]{@{}c@{}}w/o\\ SS\end{tabular}}} & 2$\times$MA w/o FW               & 57.66          & 35.00             & 49.91          & 36.82            \\
\multicolumn{1}{c|}{}                                                                    & 4$\times$MA w/o FW              & 55.03          & 33.05          & 47.63          & 36.79            \\
\multicolumn{1}{c|}{}                                                                    & 2$\times$FW+2$\times$MA & \textbf{59.62} & \textbf{36.26} & \textbf{50.91} & \textbf{37.83}   \\ \hline
\multicolumn{1}{c|}{\multirow{2}{*}{+SS}}                                              & all w/o MLM               & 57.39          & 36.37          & 50.10           & 37.49            \\
\multicolumn{1}{c|}{}                                                                    & all                       & \textbf{61.24} & \textbf{38.04} & \textbf{52.14} & \textbf{38.57}   \\ \hline
\end{tabular}%
}
\caption{Ablation study (\%) of MLM. FW and SS denote FW-MESM and SS-MESM, w/o means without.}
\label{tab:ablation_FW-MESM}
\vspace{-4pt}
\end{table}

We compare our MESM with the following state-of-the-art methods: MCN~\cite{anne2017localizing}, CAL~\cite{escorcia2019temporal}, XML~\cite{lei2020tvr}, 2D-TAN~\cite{zhang2020learning}, VSLNet~\cite{zhang2020span}, LGI~\cite{mun2020local}, CBLN~\cite{liu2021context}, RaNet~\cite{gao2021relation},  MomentDETR~\cite{lei2021detecting}, SeqPAN~\cite{zhang2021parallel}, SMIN~\cite{wang2021structured}, VISA~\cite{li2022compositional}, MMN~\cite{wang2022negative}, UMT~\cite{liu2022umt}, QD-DETR~\cite{moon2023query}, VDI~\cite{luo2023towards}, MS-DETR~\cite{wang2023ms}. 

\noindent\textbf{Charades-STA.} As Table~\ref{tab:charades} shows, our MESM performs the best with a large margin both on the uni-modal feature extractor (VGG, GloVe) and the multi-modal extractor (C+SF, C). Compared with the strong baseline QD-DETR, our MESM obtains 3.03\% average gains in mAP$_\text{avg}$ and 4.42\% in R1@0.7 on two types of extractors. 
Though multi-modal pre-trained extractors perform better than separated ones, the modality imbalance problem still makes them hard to achieve comprehensive alignment. When our MESM models more balanced semantics, it reasonably outperforms the existing state-of-the-art methods.
Though VDI does not use SlowFast, they employ a sequence model to capture the temporal relationship based on CLIP features with spatial information in $\mathbb{R}^{L_v \times H \times W \times D}$, incurring much more computational cost. 
On the contrary, we use the global frame-level feature in $\mathbb{R}^{L_v \times D}$ and filter out some semantically irrelevant components for more balanced alignment, getting better results than VDI with much less computational cost. 

\noindent\textbf{TACoS.} Different from Charades-STA, there are much fewer but longer videos in TACoS with much more sentences within a video. 
Table~\ref{tab:tacos} shows the comparison with the state-of-the-art methods. We achieve the best in all metrics. Note MS-DETR uses multi-scale video features, which is beneficial for results, but it still suffers from the modality imbalance problem and is thus sub-optimal, we obtain 5.03\% and 2.16\% gain in R1@0.3 and R1@0.5, respectively.

\noindent\textbf{QVHighlights.} QVHighlights is a special dataset as each video only contains one sentence, which is quite challenging for our SS-MESM, making it only learn the complementary knowledge from the video modality. As shown in Table~\ref{tab:qvhighlights}, we also obtain gains in all metrics on the \textit{test} split compared with QD-DETR. When FW-MESM works normally, we analyze that the masked sentence in SS-MESM tends to perform as a prompt to learn from various videos.

\noindent\textbf{Charades-CG.} When the MESM bridges the modality gap, the model should be more generalizable to understand the relationship between videos and language queries. To validate it, we conduct experiments on the OOD settings of Charades-STA.
Table~\ref{tab:charades_cg} shows the comparison. 
For the novel-composition set, compared with VISA, we gain considerably (3.29\%) in R1@0.7, which means we can get answers with higher quality due to better modality alignment. For the novel-word set, we obtain inspiring gains (12.79\% in R1@0.7). The reason can be concluded from both FW-MESM and SS-MESM. When FW-MESM provides fine-grained discrimination ability to understand novel words from frames and other words, SS-MESM also supplements additional semantics for better understanding. Compared with QD-DETR, we also gain 5.03\% in R1@0.7, demonstrating better generalization and understanding ability.

\begin{table}[t]
\centering
\resizebox{0.45\textwidth}{!}{%
\begin{tabular}{c|cc|cc}
\hline
SS-MESM layers & R1@0.5 & R1@0.7 & mIoU  & mAP$_\text{avg}$ \\ \hline
2              & 57.10  & 35.32  & 49.75 & 36.26            \\
3              & 59.38  & 35.22  & 50.68 & 37.90            \\
4              & \textbf{61.24}  & 38.04  & \textbf{52.14} & \textbf{38.57}            \\
5              & 60.97  & \textbf{38.39}  & 52.09 & 38.50            \\
6              & 58.25  & 35.70   & 50.31 & 37.60             \\ \hline
\end{tabular}%
}
\caption{Ablation study (\%) on the layers of SS-MESM.}
\label{tab:ablation_SS}
\end{table}

\subsection{Ablation Study}

\begin{figure*}[t]
\centering
\includegraphics[width=0.85\textwidth]{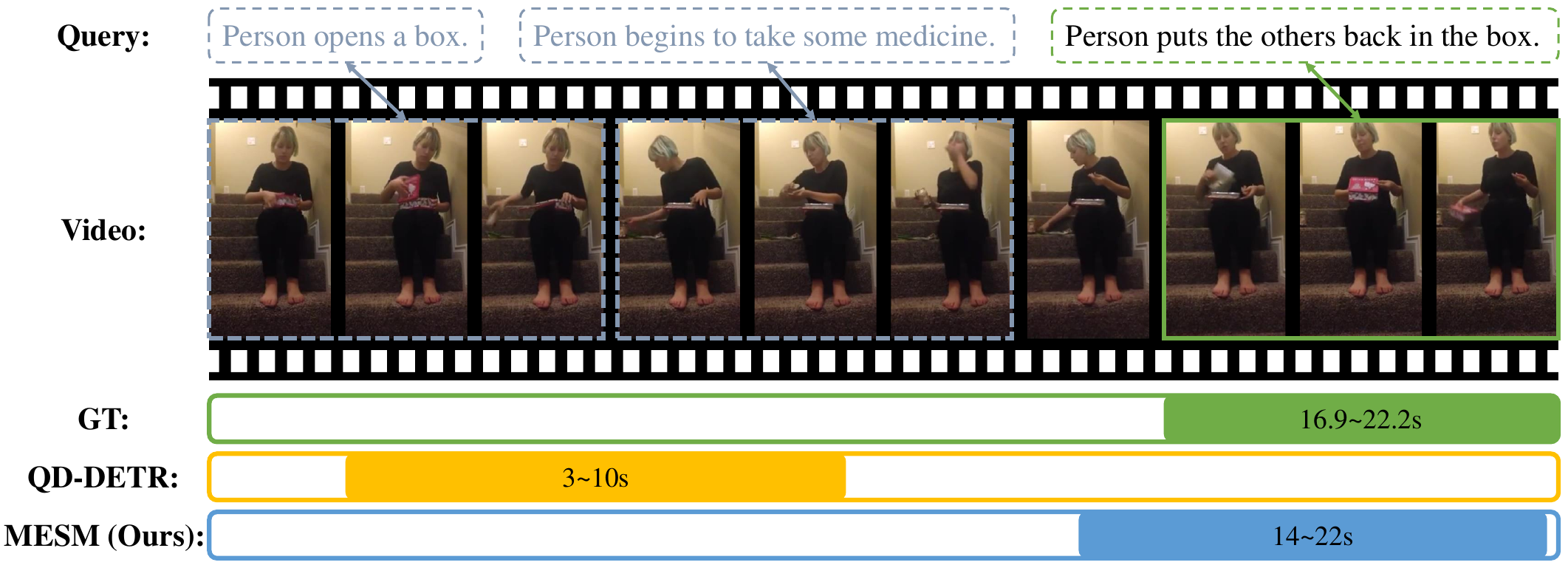}
\caption{Visualization of prediction on Charades-STA. Model needs to understand what the \textit{others} are, which is challenging.}
\label{fig_visualization}
\end{figure*}

\begin{figure}[t]
\centering
\includegraphics[width=0.4\textwidth]{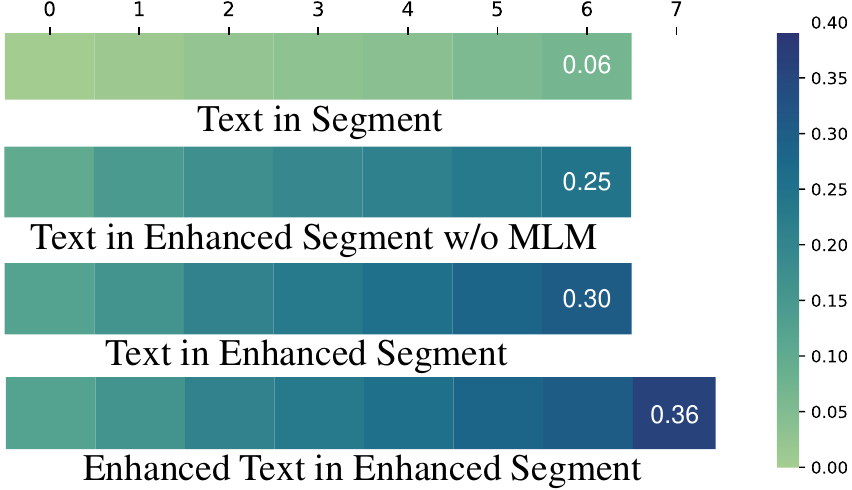}
\caption{Visualization of the level of text modality within the video modality, a reference to evaluate alignment.}
\label{fig_vis_contain}
\vspace{-4pt}
\end{figure}

To validate the effectiveness of each component, we conduct ablation studies on Charades-STA with C+SF as the video extractor and C as the text extractor. 

\noindent\textbf{Main Ablation.}
The key components of our MESM are the two different levels of semantic modeling, FW-MESM and SS-MESM, while we also add a loss function $\mathcal{L}_{enc}$ for the transformer encoder. As shown in Table~\ref{tab:ablation}, each component is beneficial for VMR and $\mathcal{L}_{enc}$ makes the most of it since it provides a supervised signal for modal-aligned features to figure out the correct segment. Without FW-MESM and SS-MESM, the framework is similar to QD-DETR, so is the performance. Based on the result of adding $\mathcal{L}_{enc}$, FW-MESM and SS-MESM achieve gains of 1\% and 1.24\% in mIoU, respectively. When we use all of them, the result comes to the best, which gains 3.04\% in R1@0.7 and 2.23\% in mIoU, demonstrating that both of them are beneficial to the results. This is because they both make the alignment more balanced and are compatible with each other.

\noindent\textbf{The Weight-Shared MLM.}
To make sure it is the weight-shared MLM works for FW-MESM, instead of the extra cross-attention layers, we conduct the ablation study as shown in Table~\ref{tab:ablation_FW-MESM}. When SS-MESM is not implemented, we set 2 layers of cross-attention (2$\times$MA) as baseline (line 1). If we simply add the layers of cross-attention to 4 without MLM (line 2), the scores drop, which may be caused by overfitting. 2 layers of FW-MESM and 2 layers of MA together achieve the best (line 3). When SS-MESM is implemented, all metrics drop with a margin without MLM. These results demonstrate the effectiveness of the MLM on the weight-shared cross-attention, instead of the extra layers.

\noindent\textbf{The Layers of SS-MESM.}
We also conduct the ablation study on the SS-MESM to figure out the suitable number of layers. As shown in Table~\ref{tab:ablation_SS}, the mIoU and mAP$_\text{avg}$ come to the best when implementing 4 layers of SS-MESM, too few layers can not provide enough power to learn the semantics while too many layers may cause overfitting. 

\subsection{Qualitative Analysis}
In Figure~\ref{fig_visualization}, we show an example of prediction. For the given query \textit{Person puts the others back in the box}, the model should understand what the \textit{others} are, which is challenging. Therefore, QD-DETR may only simply catch the words \textit{puts} and \textit{box}, then gives a wrong answer to similar scenes. When we supplement semantics and enhance both modalities, MESM understands the \textit{others} stand for the things except for \textit{medicine}, and thus gives a more accurate answer.

In Figure~\ref{fig_vis_contain}, we answer the question of how much of the text modality is contained in the video modality. We randomly select a query and calculate the subspace similarity based on the singular value decomposition~\cite{hamm2008grassmann, hu2021lora}. The calculation is implemented between the top-$i$ singular vectors of the query and all singular vectors of the segment. The similarity is quite low between the original text $F_q^\dag$ and original segment $F_v$[$l_e$:$l_s$] (line 1), and becomes much higher when it comes to $F_q^\dag$ and the enhanced segment $F_v^{enh}$[$l_e$:$l_s$] (line 2\&3) due to both the cross-modal interaction and the suppression of semantically irrelevant parts by the MLM task. The similarity between the enhanced text $F_q^{\dag enh}$ and $F_v^{enh}$[$l_e$:$l_s$] is the highest (line 4) owing to the complementary knowledge added to the textual modality. As the similarity stands for the level of containing, the results demonstrate both the enhanced video and textual modalities are more balanced than before.

\section{Conclusion}
In this paper, we address the modality imbalance problem, which means the inherently richer semantic information in the video modality than the textual modality. The imbalance makes the direct alignment sub-optimal but most methods ignore it. Therefore, we propose a novel framework MESM to tackle this problem from two levels. At the frame-word level, we enhance the video modality to filter out the redundant query-irrelevant semantics, making it more balanced with the texts. At the segment-sentence level, we enhance the textual modality to capture more segment-relevant semantics, making it more balanced with the videos. 
In the future, we aim to design a framework to model the semantic information progressively from the frame-word level to the segment-sentence level to achieve robust alignment, and we believe the issue and solution introduced in this work can provide fundamental insights to related fields.

\appendix

\section*{Acknowledgment}
This work is supported by the National Key Research and Development Program of China (2022YFB3104700), the National Nature Science Foundation of China (62121002, U23B2028, 62232006).

\bibliography{aaai24}

\end{document}